\documentclass[9pt,twocolumn,twoside]{pnas-new}
\pdfoutput=1

\templatetype{pnasresearcharticle} 

\usepackage{graphicx}

\usepackage{epstopdf}

\title{Conceptual Organization is Revealed by Consumer Activity Patterns}

\author[a,b]{Adam N. Hornsby}
\author[a]{Thomas Evans} 
\author[a]{Peter S. Riefer}
\author[a]{Rosie Prior}
\author[b,c,1]{Bradley C. Love}

\affil[a]{dunnhumby, 184 Shepherds Bush Road, London W6 7NL, United Kingdom}
\affil[b]{Department of Experimental Psychology, University College London, London WC1H 0AP, United Kingdom}
\affil[c]{The Alan Turing Institute, United Kingdom}

\leadauthor{Hornsby} 

\significancestatement{Can a hammer be a hammer if it is not used for hitting nails? Objects may become meaningful through their interactions with other objects. Using techniques developed in computational linguistics for inferring word meanings from large text-corpora, we analyzed purchasing patterns in supermarkets. These real-world activity patterns organized into topics that reflected human conceptual knowledge. Topics ranged from specific (e.g., ingredients for a stir-fry) to general (e.g., cooking from scratch). Topics tended to be goal-directed and situational, consistent with the notion that human conceptual knowledge is tailored to support action. People's age, gender, and location were predictable from their topics.These results suggest that human activity patterns reveal conceptual organization and may give rise to it.
}

\authorcontributions{T.E., P.R., R.P., A.H. and B.L.  designed research. A.H., T.E. and P.R. performed research and analyzed data. A.H. wrote the paper.}
\authordeclaration{The authors declare no conflict of interest}
\correspondingauthor{\textsuperscript{1}To whom correspondence should be addressed. E-mail: b.love@ucl.ac.uk and adam.hornsby.10@ucl.ac.uk}

\begin{abstract}
Meaning may arise from an element's role or interactions within a larger system. For example, hitting nails is more central to people's concept of a hammer than its particular material composition or other intrinsic features. Likewise, the importance of a web page may result from its links with other pages rather than solely from its content. One example of meaning arising from extrinsic relationships are approaches that extract the meaning of word concepts from co-occurrence patterns in large, text corpora. The success of these methods suggest that human activity patterns may reveal conceptual organization. However, texts do not directly reflect human activity, but instead serve a communicative function and are usually highly curated or edited to suit an audience. Here, we apply methods devised for text to a data source that directly reflects thousands of individuals' activity patterns, namely supermarket purchases. Using product co-occurrence data from nearly 1.3m shopping baskets, we trained a topic model to learn 25 high-level concepts (or \textit{topics}). These topics were found to be comprehensible and coherent by both retail experts and consumers. Topics ranged from specific (e.g., ingredients for a stir-fry) to general (e.g., cooking from scratch). Topics tended to be goal-directed and situational, consistent with the notion that human conceptual knowledge is tailored to support action. Individual differences in the topics sampled predicted basic demographic characteristics. These results suggest that human activity patterns reveal conceptual organization and may give rise to it.
\end{abstract}

\dates{This manuscript was compiled on \today}
\doi{\url{}}

\begin{document}

\verticaladjustment{-2pt}

\maketitle
\thispagestyle{firststyle}
\ifthenelse{\boolean{shortarticle}}{\ifthenelse{\boolean{singlecolumn}}{\abscontentformatted}{\abscontent}}{}

\dropcap{O}ne common view is that concepts decompose into {\em intrinsic} features or parts \cite{Plato1973}. On this view, a bird is an animal that typically has wings, feathers, a beak, and so on \cite{Rosch1975}. However, {\em extrinsic} features are also critical for how humans organize concepts and come to understand the world \cite{BarrCaplan1987}. Indeed, everyday concepts are difficult to define solely in terms of intrinsic features. For example, Wittgenstein \cite{Wittgenstein1967} asserted that the concept of game is undefinable. One might suggest that games are fun, but Russian Roulette is not fun and other activities that are fun are not games. Likewise, not all games are competitive (e.g., Ring Around the Rosie). Instead of defining game in terms of intrinsic features, one solution is to define game relationally  -- a game is simply something that is played \cite{Markman2001}. On this view, concepts are defined and become meaningful in terms of their relationships and interactions with other concepts rather than decomposing into a set of intrinsic features.

The importance of relations and interactions extends beyond abstract concepts. Many features of concrete concepts are extrinsic \cite{Jones2007}. For example, fundamental to our conception of a hammer is that it is used to hit objects, which is not an intrinsic property of hammer like its shape or material composition, but is instead a relation between a hammer and another object, typically a nail. Even for natural kinds, people commonly list extrinsic features for concepts \cite{Jones2007}, such as noting that birds eat worms. Meanings appear to update in light of extrinsic relationships. For example, people are more likely to judge a polar bear and a dog as similar after reading vignettes in which both played the same role in a relation, such as chasing some other animal \cite{Jones2007}. Likewise, merely sharing a thematic relationship, such as a man and a tie (e.g., wears), makes the linked concepts more similar \cite{Wisniewski:Bassok:1999,Jones2007}.

If concepts are defined in terms of other concepts, what moors or grounds our concepts to the physical world we inhabit \cite{Harnad1990}? One proposed solution is that some concepts are embodied \cite[for a review, see][]{Barsalou2008}. For example, the action of hammering may be grounded to related motor programs and associated perceptions, linking the body, mind, and physical world. Indeed, neuroscientific evidence has shown that comprehension of language is tightly coupled with the neural regions associated with action and perception \cite{PickeringGarrod2013}. A computational model developed by Mitchell et al. \cite{Mitchell2008} was able to accurately predict the neural activity elicited by a noun by considering the co-occurrence of that noun with action verbs in a large-text corpora. In effect, the action verbs, for which elicited neural activity was known, provided a grounding or bases for representing associated nouns.


These corpora models, such as \textit{Latent Semantic Analysis}, use the co-occurence of words within some context (e.g., a document) to learn lower dimensional, vector representations of word concepts \cite{Deerwester1990}. Like the reviewed psychological research \cite{Jones2007}, words need not directly co-occur with one another to become more similar, but need only occur in similar contexts. Although LSA has enjoyed numerous successes, cases in which its representations diverge with those of humans has prompted further model development \cite{Wandmacher2008}.

One subsequent proposal, \textit{Latent Dirichlet Allocation (LDA)}, is a probabilistic approach in which documents are generated according to a mixture of probabilities over latent themes or topics \cite[e.g.,][]{Blei2003}. For example, LDA may find that the words `Parliament' and `Prime Minister' have a high probability of belonging to the same topic (e.g., `politics'). A passage about the Prime Minister visiting the Houses of Parliament would make this politics topic highly probable, though other topics would also be somewhat likely, such a topic related to tourism (Big Ben is part of the Houses of Parliament). 

The representations learned by topic models appear similar to the concepts that people use \cite{Griffiths2007, Andrews2009}. For example, topic modeling can predict subsequent words in a sentence, disambiguate word meanings, and extract the gist of a sentence \cite{Griffiths2007}. Related techniques find that word meanings extracted for text corpora reflect back that society's gender stereotypes \cite{Bolukbasi2016}. These successes provide credence to the idea that human concepts are heavily influenced by their extrinsic roles and relationships. It may be through patterning and interaction that concepts gain meaning, as opposed to decomposing into intrinsic features.  

Despite these successes, corpora analysis of natural language is not ideally suited to evaluate the proposal that human concepts become meaningful and organize around extrinsic relations. After all, language is primarily concerned with effective communication of relevant information \cite{Grice1975}, rather than providing a faithful record of object interactions. For example, in waiting to cross the street with a companion, one would never verbalize that the passing car drives on the road. Written language is also heavily curated. For example, journalists adhere to particular guidelines and aim to report on stories of interest to their readership. Thus, data collected from either spoken or written language is arguably unrepresentative of true activity patterns.

\begin{figure}[ht]
\centering
\includegraphics[scale=1.0]{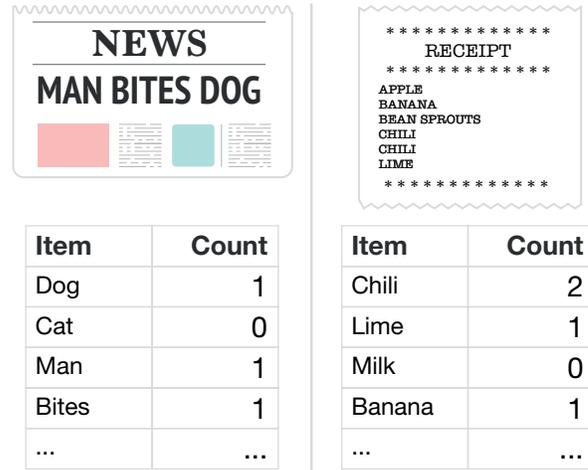}
\caption{The input in a corpora analysis is typically  item counts (i.e., word counts) within some context (e.g., a sentence or document). Analogously, products (akin to words) are organized into baskets (akin to sentences). One advantage of applying these analysis techniques to baskets is that, unlike natural language, meaning is unaffected by item order.}
\label{fig:words_products}
\end{figure}

The ideal data set for topic modeling would contain unfiltered and unedited information that directly pertains to human activity patterns. Consumer shopping is one such test bed. Retail data are collected from consumers as they purchase products together in the same basket, analogous to how words group together in the same document (see \ref{fig:words_products}). Unlike natural language, grocery transaction data are not carefully curated with a particular audience in mind. Moreover, there are no editors controlling the final assortment of items in a basket (i.e., document). Unlike purchasing data collected from many thousands of consumers, text found in newspapers is typically written by a select cadre of journalists, thereby biasing the data towards the experience of a small minority. Therefore, one could argue that consumer purchases are more representative of activity patterns in-the-wild and therefore serve as a more appropriate test bed for evaluating models and theories of human semantic representation. In effect, working with a such a dataset would provide a real-world, large-scale test of theories of meaning that were hitherto only possible to explore in laboratory studies under limited and artificial conditions.

In addition to being more representative of people's activity, consumer purchasing data better suits the mathematical assumptions of topic models than does natural language. Indeed, natural language researchers typically go to great lengths to pre-process their data. For example, researchers typically must remove function or `stop' words that have little semantic meaning (such as \textit{the, of, and}). They may also `stem' words to remove prefixes and suffixes of words that have similar semantic meaning (e.g., eat vs. eating). Moreover, the order of words in sentences can also make a big difference to sentence meaning (e.g., ``dog bites man'' vs. ``man bites dog''). However,  topic models typically ignore word order, instead preferring to consider language as a ``bag-of-words'' \cite{Huang2015}. In contrast, for retail data captured in-store, there is no inherent order for products within a basket, nor a need to heavily pre-process the data.

\begin{figure*}[ht]
\centering
\includegraphics[scale=1.0]{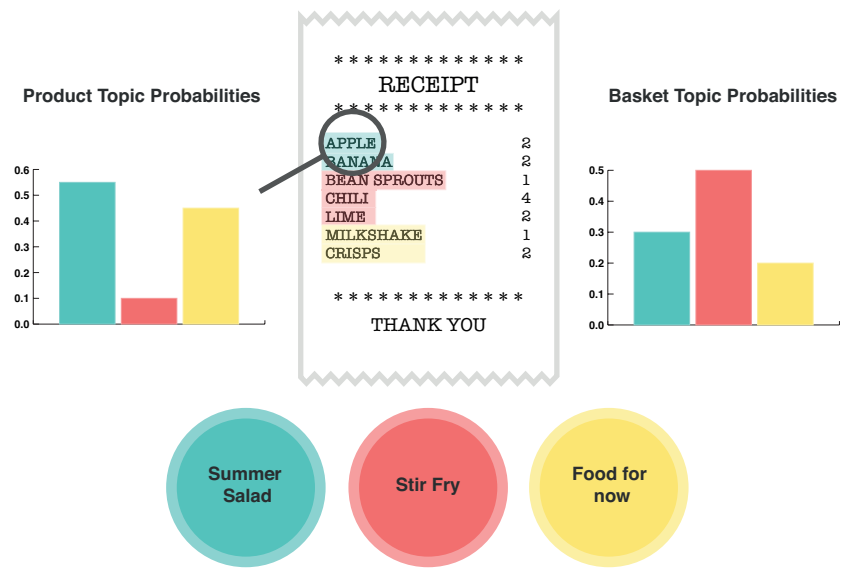}
\caption{Latent Dirichlet Allocation (LDA) uncovers the higher-level product topics that can be viewed as generating the observed baskets purchased by consumers. LDA's fit is driven by the co-occurrence pattern of products within baskets. In the solution, each product has a probability of occurring within each topic (shown on the left for apple). Each basket is generated by a mixture of probabilities over the topics (shown on the right for this basket).}
\label{fig:basket-generation}
\end{figure*}

If people's thematic organization of concepts arises through their interaction with the environment, then it should be possible for a topic model to recover relevant representations of these through consumer purchasing patterns, as shown in Figure \ref{fig:basket-generation}. We tested this possibility using a large, anonymized dataset of 1,252,963 shopping baskets and 5,753 unique products, supplied by one of the UK's largest supermarket retailers. After optimizing an LDA solution using fit statistics and checking for convergence\footnote{More detail about the model fitting procedure can be found in the methods section}, we labelled the 25 topics recovered by the model. 

To foreshadow, we confirmed the psychological reality of these topics in two human studies, one with judgments from retail experts and another involving typical consumers. The latter study also suggested that an individual's shopping experience shaped their conceptual organization of the products. In support of this assertion, the rate at which an individual sampled topics (based on recent shopping history) predicted the individual's age, gender, and geographic region. Finally, topics tied to a season varied sensibly in their prevalence over the calendar year (e.g., the Christmas topic was most prevalent in December). Overall, these results suggest that conceptual organization may arise from people's direct interactions with objects, and, conversely, that patterns of real-world behaviour may reveal a great deal about an individual and their goals.

\section*{Results}

The topics recovered were coherent and readily labeled by the authors. They tended to be organized along activity patterns and goals, ranging from specific (e.g., \textit{Stir Fry}) to general in scope (e.g., \textit{Cooking from scratch})\footnote{A complete list of the 25 topics and process used to label them is available in the supplemental}. A subset of 10 topics was considered in the empirical studies of retail experts and typical consumers.

\subsection{Evaluating topic labels with retail experts}
\begin{figure}[ht]
\centering
\includegraphics[scale=1.0]{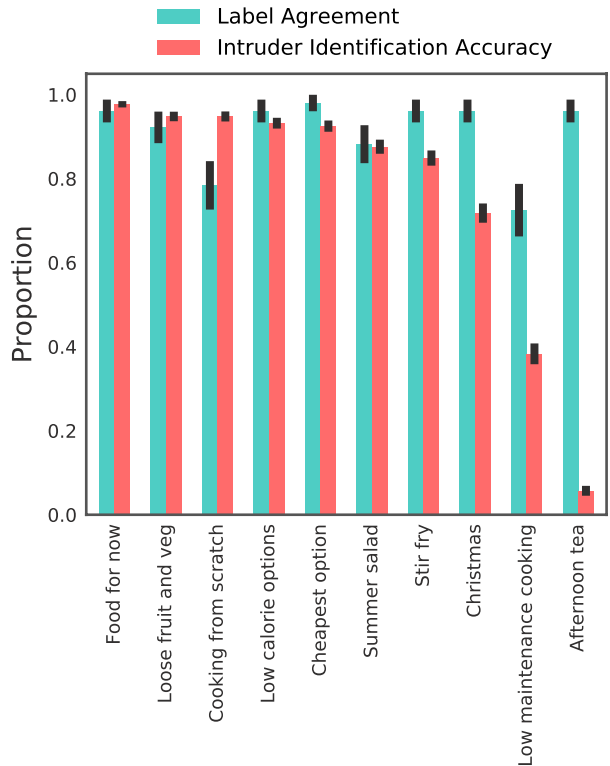}
\caption{Proportion correct with standard error bars for the study on label agreement involving retail experts and the intruder study involving typical consumers. All proportions were significantly different ($p < .001$) than chance levels, 25.00\% (1 of 4) and 16.67\% (1 of 6), respectively.}
\label{fig:both_studies}
\end{figure}

When asked to select the appropriate topic label for a group of products from a list of four possible labels, 92.8\% of the 51 retail industry experts selected the same topic label as was originally proposed by the authors. Figure \ref{fig:both_studies}  shows the proportion of times participants agreed with the originally proposed topic label for each topic. 

This high-level of accuracy from a group of experts, who were naive to our research program, indicates that the topics recovered from consumer activity patterns are meaningful. Disagreement regarding topic labels was primarily driven by conceptually similar topics (further details of this are available in the supplemental). For example, the most common error in labeling the \textit{summer salad} topic was \textit{cooking from scratch}. These errors are reasonable and are also consistent with the notion that baskets are generated by a mix of topics, as opposed to a single topic (see Figure \ref{fig:basket-generation}).

\subsection{Evaluating topic coherency with typical consumers}

We evaluated whether typical consumers could identify an intruder product from another topic among a number of products from the same topic. The 5 products from the same topic and the 1 intruder product were all highly ranked within their respective topic\footnote{More information about the ranking procedure used is available in the supplemental}. Of the 3840  British consumers surveyed, 76.0\% were able to correctly identify the intruder product. Figure \ref{fig:both_studies} shows accuracy by topic.

One topic stands out for its below chance level of performance, \textit{afternoon tea}.  Participants were most likely (51.7\% of the time) to incorrectly suggest that `mint humbugs' was the intruder. One possibility for this poor classification accuracy is that participants did not have enough context to interpret them correctly. In the \textit{afternoon tea} topic, the top 5 items were predominantly fresh and `staple' foods (e.g., Milk, Bananas, Danish sliced white bread). Thus, seeing a packet of sweets (i.e., `humbugs') among this fresh food may have appeared unusual. An analogous issue arises with the \textit{low maintenance cooking} topic. Each topic is a probability distribution over thousands of products, so perhaps it is not surprising that a small sample of products could be ambiguous.

Another possibility that is more core to our theory is that individual differences in experience may help explain some of these confusions. For example, the poor classification of the \textit{afternoon tea} topic may have been driven by the fact that most British people no longer regularly engage in this ritualistic activity. If experience shapes people's mental concepts, then we would expect representations of certain products to vary between demographics. Supporting this view, consumers from Northern Ireland had an average probability for the \textit{Northern Ireland} topic $7.5 \times$ higher than the average across all regions. The fact that the model was able to recover such strong regional differences in consumers suggests that it should be sensitive to other individual differences in people's experience of the world.

\subsection{Classifying individual consumers by their experienced topics}

The previous results hinted that individual differences in knowledge organization may arise from differences in product interactions. More ambitiously, differences in how often people experience topics may predict basic demographic characteristics. To test this assertion, we used logistic regression to predict (5-fold cross-validation) self-reported age \footnote{Discretized into 18-29, 30-44, 45-59 and 60+}, region \footnote{Binarized into England vs. regional (i.e., Scotland, Wales and Northern Ireland)} and gender using consumers' mean LDA probabilities over baskets as the predictors. The models were able to predict age with an accuracy of 48.51\%, region with an accuracy of 58.34\% and gender with an accuracy of 57.17\%, considerably higher than the guessing baselines of 36.86\%, 50.00\%, and 50.00\%, respectively. 

\subsection{Seasonal trends in topics}

In addition to experiences of topics varying across individuals, prevalence of topics should vary over the calendar year. We identified 4 topics that were likely to have a high seasonal popularity (\textit{summer fruits}, \textit{summer salad}, \textit{Christmas} and \textit{low calorie options}) and 4 `staple-food' topics that we believed unlikely to vary as much over the year (\textit{loose fruit and veg}, \textit{Northern Ireland}, \textit{quick to prepare meals} and \textit{food for now}).

\begin{figure}[ht]
\centering
\includegraphics[scale=1.0]{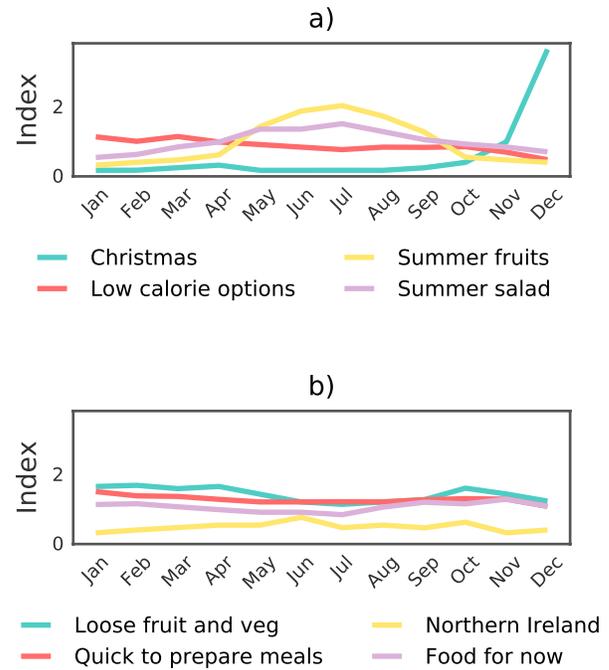}
\caption{Topic prevalence varies by season. The proportion of baskets with a given topic label in each month of 2014, divided by the monthly mean average across all topics (i.e., index), is shown. a) Topics that should be seasonal peak at the expected time, such December for the Christmas topic. b) In contrast, topics for staple products vary less in prevalence over time.}
\label{fig:seasonal_trends}
\end{figure}

Figure \ref{fig:seasonal_trends} shows the proportion of times baskets associated with a given topic occurred each month in 2014 in one of the UK's largest retailers. In line with our hypotheses, \textit{summer fruits} and \textit{summer salad} peaked in popularity during the summer months. Contrasting, baskets labelled with the \textit{Christmas} topic peaked in popularity during December and the surrounding winter months. \textit{Low calorie options} appeared to peak in January and steadily decline to its lowest level of popularity in December. The `staple' topics shown in Figure \ref{fig:seasonal_trends}b appeared to vary considerably less over the year compared to the more seasonal topics. These results give further credence to the proposed topic labels and illuminate some seasonal variations in behavioural patterns that likely reflect time-dependent characteristics of people's thematic representations. Similarly intuitive patterns were shown to occur during different days of the week, which are reported in more detail in the supplemental.

\section{Discussion}

Rather than meaning residing solely in intrinsic features, conceptual knowledge may arise from object interactions. For example, the meaning of hammer is more closely tied to hitting nails than to its particular material composition. Support for the notion that meaning arises from a web of interactions comes from laboratory studies demonstrating that object interactions alter how people conceptualize objects \cite{Jones2007} and from large-scale corpora analysis of text (e.g., newspaper articles) that extract meaning from word co-occurrence patterns. However, none of these previous investigations involve people engaging in unfiltered, goal-driven, real-world interactions with objects. Under such conditions, can meaningful conceptual organization be recovered from human activity patterns?

We tested this possibility by considering the shopping patterns of thousands of UK consumers. Using LDA, we found that the pattern of consumer purchases was highly revealing of people's conceptual organization of these products. Topics ranged from specific and goal-driven (e.g., ingredients for a stir-fry) to very general (e.g., cooking from scratch). Interestingly, the topics tended to be  goal-directed and situational, which is consistent with the notion that human conceptual knowledge is tailored to support action. The situational nature of certain topics was reflected in their increasing prevalence during certain times of the year, such as the \textit{Christmas} topic in December and the \textit{Summer Salad} topic in the Summer.

The psychological reality of the 25 LDA topics we found was confirmed by two studies, one involving retail experts and one involving everyday consumers. The experts, who were blind to the purposes of this research, agreed with our labeling of the topics. The novices were able to identify an intruder product among an array of products from the same topic. These results indicate that the topics uncovered by human activity patterns are both comprehensible and coherent. 

If meaning and real-world behavior are linked, then individual differences in experiences should be reflected in differences in conceptual organization. In support of this conjecture, topic prevalence varied across geographic regions. In our study of everyday consumers, poor performance for the topic \textit{afternoon tea} may reflect that today's typical British consumer differs from past caricatures. Consistent with the idea that different types of people will have different topic experiences, we were able to predict basic demographic information about consumers from their topics mix (i.e., which topics best characterized their purchasing behavior). One avenue for future research is to develop, apply, and evaluate topic models in which individuals organize into higher-level groups that can vary in terms of topic prevalence or even topic composition.

One interesting question is whether shopping activity changes conceptual organization or conceptual organization drives shopping behaviour. Our results cannot definitely answer this question, but the likely answer is that the influence is bidirectional, much like how choices follow from preferences and preferences to a degree follow from choices \cite{Riefer2017}. For example, having a concept like \textit{stir-fry} should cause certain items to be purchased together to fulfill the common goal. Likewise, ingredients in the same dish may come to be viewed as more similar over time, consistent with laboratory studies that find that linking objects makes them more similar \cite{Jones2007}. One practical ramification is that recommender systems \cite{Vasile2016, Christidis2010} using techniques related to our own may themselves shape conceptual organization.

What is clear is that conceptual organization is deeply tied to extrinsic relationships and that meaning can be seen as a byproduct of an element's role within a larger system or web. Indeed, the insight behind Google's PageRank algorithm is that web pages should be prioritized to the extent that they are central within a link graph \cite{Pageetal98}. Prior to PageRank, the exact same algorithm was developed in Psychology to explain why certain features of concepts are more central than others within a concept web \cite{LoveSloman1995}. Whether the system is human or artificial or the domain involves natural language or shopping behavior, meaning can be inferred, and perhaps arises, from relations among elements embedded within a larger system.


\matmethods{

\subsection{Topic Model}

\subsubsection{Data}

The topic model was developed on a random 0.1\% sample of all grocery transactions that occurred in 2014 in one the UK's largest supermarket chains. The transactions were filtered such that only relatively popular products selling $> 50,000$ units annually were kept. Moreover, data was filtered such that only large baskets containing $\geq 20$ items were kept. Filtering was performed to ensure that LDA would have enough observations to learn meaningful topics. This is typical in LDA modelling \cite[see][]{Yan2013} and is performed by the original LDA authors \cite{Blei2003}. After filtering, the final dataset contained 1,253,183 unique baskets and 5,753 unique products.

Note that --- unlike traditional uses of LDA in NLP --- we did not remove commonly occurring items from documents (i.e., `stop words'). Whilst natural language may contain `stop words' (i.e., common words with little semantic meaning such as `the'), we did not believe grocery transactions to suffer from the same problem. In the retail case, purchasing popular products, such as milk, bananas and bread, may be informative, perhaps indicating that the consumer is stocking essential items.

The basket data was fully anonymised for general research purposes so as to not be personally identifiable.

\subsubsection{Model fit}

In our experiments we applied Latent Dirichlet Allocation (LDA) to the data, using the machine learning library in Spark 1.6.0 \cite{Spark}.  We conducted a range of experiments to identify the optimal set of hyperparameters (including the number of topics $k$) and in each case monitored the training and test log-perplexity to ensure model convergence and generalization, respectively (see supplemental for further details).

The LDA solution with the lowest log-perplexity on held out data (i.e., best generalization) had 25 topics. Models were trained for a maximum of 500 epochs, used the Online Variational Bayes optimization algorithm with an $\alpha = 0.1$. The remaining hyperparameters were set to the package defaults.


\subsection{Retail experts study}

\subsubsection{Participants}

Participants were recruited internally within the UK headquarters of dunnhumby (\url{www.dunnhumby.com}), a customer marketing company with over 29 years of experience working with grocery retailers and fast moving consumer goods (FMCG) brands. Employees were asked to participate via the company intranet and were not remunerated. Fifty-one participated in the study. Participants had a wide range of roles within the business, including data analysts, category experts, company directors and client leads. Of these, 56.86\% were male. Participants were surveyed in early December 2016 and were blind to the purposes of this study.

\subsubsection{Materials}

The study was hosted on an internal company server. Participants accessed the study via their web browsers and answered questions by clicking on the appropriate radio button with their cursor. 

In each trial, participants were shown 10 product images and accompanying product descriptions from a single topic. The displayed products were the 10 with the highest \textit{relevance}. Product descriptions and images were downloaded from the retailer's website in late November 2016.

\subsubsection{Design}

All participants were asked to label the same 10 topics in a random order. The dependent measure was the proportion of times that participants selected the topic label originally proposed during the model development phase. This proportion was then compared against a random baseline, to check whether participants were responding non-randomly.

It was not feasible to survey participants about all 25 topics in the final LDA solution given constraints on employee time. Therefore, 10 topics were chosen from the original 25 to include in the survey.

\subsubsection{Procedure}

Participants were asked to label a group of products for 10 separate topics. The order of the presented topics was randomized. In each trial, participants saw 10 products from a certain topic and were asked to choose from one of four labels. One of the four labels was the `target' label proposed by the authors whereas the other three were randomly selected from the remaining nine topics. The 10 products were the most \textit{relevant} ones from within that topic. The process used to determine item relevance within a topic is described in the supplemental.

\subsection{Consumer study}

\subsubsection{Participants}

Participants were recruited using dunnhumby's consumer survey panel; Shopper Thoughts (\url{https://shopperthoughts.com/}). Participants completed the survey as part of a larger, monthly survey for 50 card loyalty points.
The final sample consisted of 3,840 participants, of which 59.47\% were female.
The modal age group was 55-59 ($n$ = 501) and 724 participants did not disclose their age. 
All participants were from England, Scotland or Wales with the majority of respondents based in central England ($n$ = 946). Participants were surveyed during March 2017.

\subsubsection{Materials}

The study was accessible via the web, after logging in to the survey platform. Participants responded to the survey by clicking on a radio button next to the picture and product description of the item they believed to be the intruder.

For each trial, participants were shown 6 product images. Five were the most relevant from within a topic and the other `intruder' was the most relevant product from a randomly selected alternative topic. Product descriptions and images were the same as those used during Study 1.

\subsubsection{Procedure}

Participants were first informed that the purpose of the study was to help retailers group together products found in the supermarket. They were then informed about the study's procedure. After agreeing to participate, the sole trial started.

Participants were shown six images of products alongside product descriptions and asked to select the item that didn't belong to the group by clicking on a radio button underneath the appropriate image. Following their choice, participants were debriefed about the purpose of the research.

\subsubsection{Design}

The dependent variable was the proportion of times that participants identified the intruder product. This proportion was then compared against a random baseline to assess whether participants were able to identify intruders significantly above chance levels.

Participants each completed one trial in which they were asked to identify the intruding product. Participants were randomly selected to see one of the 10 topics also used in the retail experts study. This ensured that comparisons between the two related studies were consistent.

\subsection{Predicting individual differences}

To demonstrate that the LDA solution could predict meaningful attributes about customers, we built three Regularised Logistic Regression models. In particular, we averaged the LDA probabilities from each basket at the customer level and then used those mean probabilities to predict customer's self-reported age band, region and gender. These self-reported values had been gathered by the marketing panel used for the consumer study during the last 3 years. The final dataset consisted of data from 30,233 consumers. Age was discretized in to 4 buckets; 18-29, 30-44, 45-59 and 60+. The age model was therefore evaluated in terms of accuracy. Region was collapsed into two buckets; England and regional (i.e., Scotland, Wales and Northern Ireland). The gender and region models were therefore measured in terms of the Area Under the ROC curve (AUC). The final dataset consisted of 28,122 customers. To find the best performing model, we performed a grid-search between $\lambda$ values of 0.1 to 1.0. We determined the best model to be the one that had the highest performance over 5 cross-validation folds. Baselines were calculated by predicting the majority class in each fold.

}

\showmatmethods{} 

\clearpage

\acknow{We thank Gareth Brown, Olivia Guest, Sebastian Bobadilla Suarez, Brett Roads and Kurt Braunlich for their help and feedback. A. N. H. is supported by dunnhumby and the 1851 Royal Commission. B.C.L. is supported by Leverhulme Trust RPG-2014-075, Wellcome Trust Senior Investigator Award WT106931MA, and National Institute of Child Health and Human Development Grant 1P01HD080679.}

\showacknow{} 


\bibliography{lda_bib}

\end{document}


\section{Supplementary Information}

\subsection{Latent Dirichlet Allocation}

For this project, we used a topic model known as Latent Dirichlet Allocation (LDA) \cite{Blei2003}. LDA is a generative probabilistic model that groups data into $K$ unobserved topics. In the case of this project, baskets are represented as random mixtures over unobservable topics. Topics are then characterized as a mixture over $N$ distinct products and $D$ baskets. The generative process used by LDA can be described as follows:

\begin{enumerate}
    \item For each topic $k \in \{1, ..., K\}$:
    \begin{itemize}
    \item Choose a distribution over products $\phi_k \sim Dir(\beta)$.
    \end{itemize}
    \item For each basket $d \in \{1, ..., D\}$ in the collection $\boldsymbol{c}$:
    \begin{itemize} 
    \item Generate a vector of topic probabilities: $
    \theta_d \sim Dir(\alpha)
    $
    \item For each product $w_{d,n}$ in basket $d$:
    \begin{itemize}
        \item Generate a topic assignment: 
        $z_{d,n} \sim Multinomial(\theta_d)$
        \item Draw a product $w_{d,n} \sim Multinomial(\phi_{z_{d,n}})$
    \end{itemize}
    \end{itemize}
\end{enumerate}

Where $\beta$ and $\alpha$ are hyperparameters that determine the concentration of the Dirichlet prior placed on the topics' distribution over products $\phi$ and the baskets' distribution over topics $\theta$, respectively. The latent variables $\phi_k$ and $\theta_d$ can then be inferred using an iterative learning algorithm, such as expectation maximization \cite{Blei2003}.

\subsubsection{Topic inspection and labelling}

\begin{table}[ht]
\centering
\caption{The labels given to each of the 25 topics. Size is defined as the number of products that had the highest probability of belonging to the respective topic over the total number of products in the corpus. Asterisks indicate that they were surveyed in studies A and B.}
\label{table:final_topics}
\begin{tabular}{ll}
Topic label & Size \\
\midrule
Loose fruit and veg * &	18.8\% \\
Young children's shop &	18.0\% \\
Own brand shop &	17.5\% \\
Cooking from scratch * &	16.9\% \\
Snacks &	16.4\% \\
Cheapest option * &	15.4\% \\
Home baking &	14.9\% \\
Exotic cooking from scratch &	14.5\% \\
Afternoon tea * &	14.5\% \\
Quick to prepare meals &	13.9\% \\
Branded store cupboard &	13.9\% \\
Summer salad * &	13.2\% \\
Summer fruits &	12.6\% \\
Low maintenance cooking * &	12.5\% \\
Low calorie options * &	11.3\% \\
Party snacks &	9.3\% \\
Christmas * &	7.3\% \\
Northern Ireland &	6.3\% \\
Delisted Products &	5.8\% \\
Cat lover &	1.3\% \\
Stir fry * &	1.1\% \\
Own brand family party &	1.0\% \\
Food for now * &	1.0\% \\
Eating from tins &	0.5\% \\
Desserts &	0.2\% \\
\bottomrule
\end{tabular}
\label{table:topics}
\end{table}

For each experiment, we calculated model \textit{perplexity} on the training set and a held out test set. The perplexity of a model on that collection is defined as:





$$
Perplexity(c) = exp \bigg \{{\frac{- \sum_d log \:  p(w_d)}{\sum_d N_d} \bigg \}}
$$

Where $p(w_d)$ is the probability of a product $w$ in a basket $d$ and $N_d$ is the number of products in a basket. Perplexity can be useful to monitor during model training to ensure that the algorithm is converging on the training set and generalizing to unseen data. However, some have argued that perplexity and human interpretation are uncorrelated or even negatively correlated \cite{Chang2009}.

Given that interpretability is a fundamental goal of this research, we decided to test it more directly. In particular, we tested to see whether the topics were interpretable to humans and could identify known patterns in historic purchasing data. We now discuss this in more detail.

\subsubsection{Calculating product relevancy for a topic}

One conventional approach to interpreting topic models is to rank items (i.e. products) within a topic and manually inspect those with the highest probabilities. One can look for similarities between items with high probabilities to understand whether there is a key theme that binds them together.

A major issue with the traditional approach to interpreting topic models is that the probability of an item given a topic $\phi_{kw}$ can be biased positively in favour of more frequently occurring items within the corpus \cite{Taddy2012}. The validity of this traditional approach was therefore significantly limited, particularly given that we did not filter high-frequency products (i.e. stop items) from our dataset.

To overcome issues with biased item probabilities, we explored two additional measures for determining the pertinence of items within a topic; \emph{lift} and \emph{relevance}. 

The lift measure \cite[first described by][]{Taddy2012} is defined as:


\begin{equation}
    lift(w,k)  = \frac{p(w_k)}{p_w}
\label{eq:lift}
\end{equation} 

Lift is therefore a ratio of an item $w$'s probability within a topic $k$ (i.e., $p(w_k)$) to its marginal probability across the corpus ${p_w}$. Whilst this helps to diminish the impact of overall token frequency, some have argued that the measure is noisy \cite{Sievert2014}. In particular, it can give overly high rankings to items that occur very rarely within the corpus. Indeed --- during our manual inspection of the topics --- we found this to be the case, limiting our ability to interpret the topics' meanings.

To overcome this problem, Sievert and Shirley \cite{Sievert2014} proposed the \textit{relevance} metric, which is defined as:

\begin{equation}
    r(w, k \vert \lambda) = \lambda \:  log \: p(w_k) + (1 - \lambda) \: log \left (\frac{p(w_k)}{p_w}\right)
\label{eq:relevance}
\end{equation}

where $\lambda$ is a free parameter that determines the weight given to the item $w$'s probability within a topic $k$ relative to its lift (measured on a log scale). If one sets $\lambda = 1$ then $r(w, k \vert \lambda) = log \: p(w_k)$. Alternatively, if $\lambda = 0$ then $r(w, k \vert \lambda) = log(lift(w,k))$. Thus, the benefit of this metric over lift is that it's possible to blend the probability of an item given a topic with lift. The metric's authors recommend using $\lambda = 0.6$ to maximize human interpretability, which is the value that we kept throughout all analyses \cite{Sievert2014}.

\begin{figure}[ht]
\centering
\includegraphics[scale=1.0]{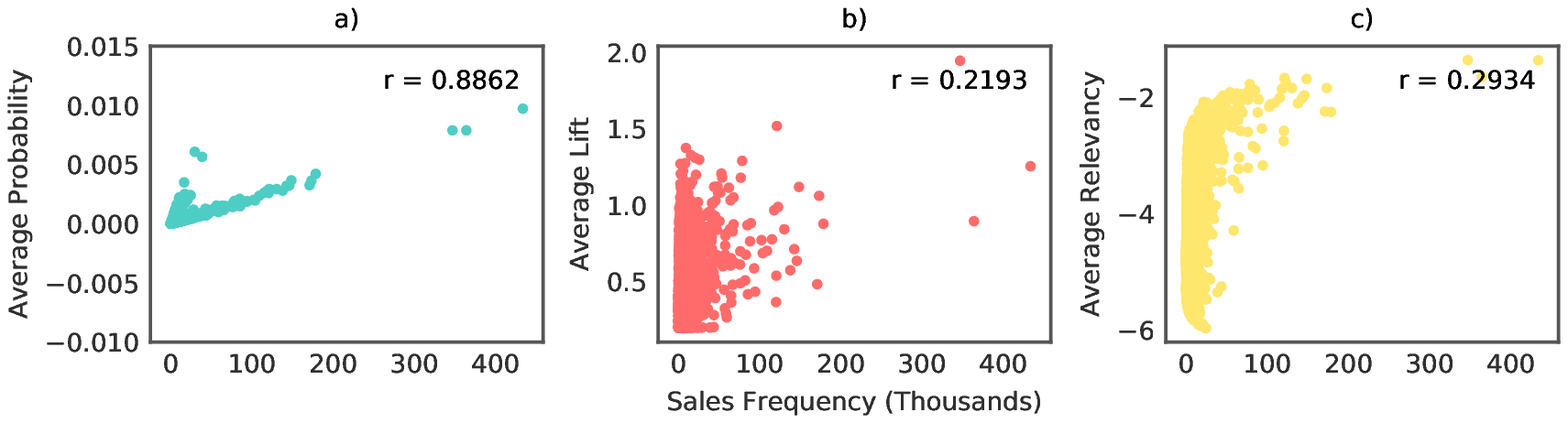}
\caption{The relationship between item frequency within the corpus and a) item-topic probability, b) lift and c) relevancy}
\label{fig:relevancy_scatter}
\end{figure}

The data displayed in Figure \ref{fig:relevancy_scatter} plot the relationship between item corpus frequency and each respective metric. As discussed, results indicate a strong correlation between item probabilities and frequency ($r = 0.8862$). The lift metric ($r = 0.2193$) and relevance metric ($r = 0.2934$) considerably reduce this correlation.

After manually inspecting the topics with each of the three measures, we agreed that relevance provided the best measure of item salience within a topic. We therefore used this to help determine topic names.

\subsubsection{Initial topic labelling}

When labelling the topics, the authors inspected the relevancy scores of each item within each topic, considered the most relevant items. Table \ref{table:final_topics} shows the topic labels along with the relative size of each topic within the corpus.

\subsection{Label confusion by retail experts}

\begin{table}[ht]
\centering
\caption{A summary of mislabelling errors made by retail experts during the topic labelling task (*** indicates a proportion significantly different ($p < .001$) from the random baseline of 25.00\% (1 of 4))}
\begin{tabular}{lrrrr}
Topic label &   Proportion correct &      Most confused topic & Most confused frequency \\
\midrule
         Cheapest option &   0.98 *** &                          &                   0 \\
            Food for now &  0.961 *** &  Low maintenance cooking &                   2 \\
                Stir fry &  0.961 *** &                          &                   0 \\
     Low calorie options &  0.961 *** &                          &                   0 \\
               Christmas &  0.961 *** &             Food for now &                   2 \\
           Afternoon tea &  0.961 *** &          Cheapest option &                   2 \\
     Loose fruit and veg &  0.922 *** &     Cooking from scratch &                   3 \\
            Summer salad &  0.882 *** &     Cooking from scratch &                   3 \\
    Cooking from scratch &  0.784 *** &      Loose fruit and veg &                   8 \\
 Low maintenance cooking &  0.725 *** &             Food for now &                   7 \\
\bottomrule
\end{tabular}
\label{tab:study1_tresults}
\end{table}

In the retail expert study, errors in labeling appeared sensible.
 Namely, the most popular alternative labels tended to be related to the original topic (see Table \ref{tab:study1_tresults}). For example, the most popular alternative label to the \textit{cooking from scratch} table was \textit{loose fruit and veg}; both topic labels pertain to ingredients that need to be prepared before consumption. 

\subsection{Topics by day of week}

In addition to monthly trends, the proposed topic labels are also indicative of different weekly trends in purchasing habits. In particular, we hypothesized that topics indicative of longer preparation times (e.g. \textit{loose fruit and veg}) or a special weekend occasion (e.g. \textit{afternoon tea}) would be more likely to occur on or just before the weekend. Contrasting, we hypothesized that topics indicative of impulse purchasing (e.g. \textit{food for now}) or stocking up for the long-term (e.g. \textit{branded store cupboard}) would not vary much across the week. 

\begin{figure}[ht]
\centering
\includegraphics[scale=0.7]{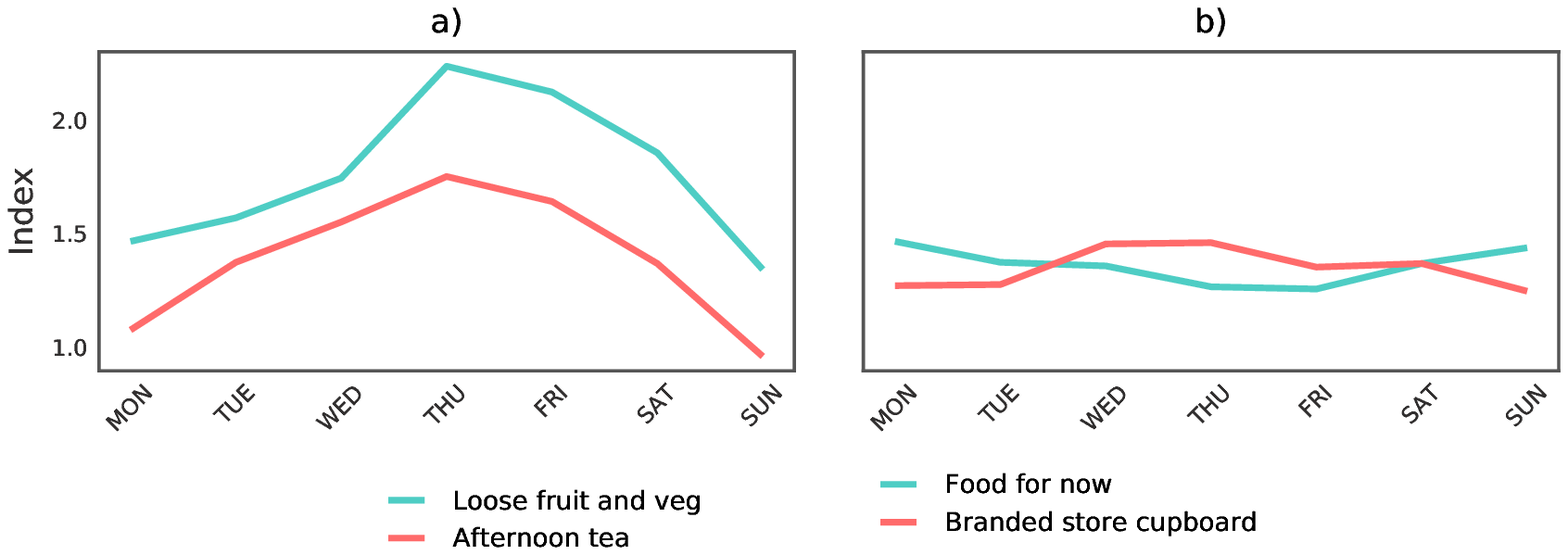}
\caption{The proportion of baskets with a given topic label on each day of the week, divided by the weekly mean average across all topics. Plot a) shows that food requiring longer preparation times (i.e. \textit{loose fruit and veg}) or eaten specifically during weekend occasions (i.e. \textit{afternoon tea}) are more likely to be bought on Thursday or Friday. Plot b) indicates that impulse purchases (i.e. \textit{food for now}) or food that tends to be stored away (i.e. \textit{branded store cupboard}) does not vary in popularity over the week.}
\label{fig:weekly_trends}
\end{figure}

\bibliographystyle{apacite}
\bibliography{supp}